\documentclass[sigconf,authorversion,nonacm]{acmart}
\usepackage{amsmath,amsfonts} 
\usepackage{graphics} 
\usepackage{algorithm,algorithmic}
\usepackage{subfig} 
\usepackage{url} 
\usepackage{textcomp}
\usepackage{xcolor}
\usepackage{siunitx}
\usepackage{enumitem}
\usepackage{multirow}
\graphicspath{{figs/}}
\usepackage{bbm}
\usepackage{fancyhdr}
\usepackage{tikz}

\newcommand{\RA}[1]{\textcolor{black}{#1}}

\newcommand*\circled[1]{\tikz[baseline=(char.base)]{
 \node[shape=circle,draw,inner sep=2pt] (char) {#1};}}

\AtBeginDocument{%
  \providecommand\BibTeX{{%
  \normalfont B\kern-0.5em{\scshape i\kern-0.25em b}\kern-0.8em\TeX}}}

\begin{document}
\fancyhead{}
\title{Piracy-Resistant DNN Watermarking by Block-Wise Image Transformation with Secret Key}


\author{MaungMaung AprilPyone and Hitoshi Kiya}
\affiliation{%
 \institution{Tokyo Metropolitan University}
 \city{Tokyo}
 \country{Japan}
}

\renewcommand{\shortauthors}{Maung and Kiya}

\begin{abstract}
In this paper, we propose a novel DNN watermarking method that utilizes a learnable image transformation method with a secret key. The proposed method embeds a watermark pattern in a model by using learnable transformed images and allows us to remotely verify the ownership of the model. As a result, it is piracy-resistant, so the original watermark cannot be overwritten by a pirated watermark, and adding a new watermark decreases the model accuracy unlike most of the existing DNN watermarking methods. In addition, it does not require a special pre-defined training set or trigger set. We empirically evaluated the proposed method on the CIFAR-10 dataset. The results show that it was resilient against fine-tuning and pruning attacks while maintaining a high watermark-detection accuracy.
\end{abstract}



\maketitle

\section{Introduction}
Deep learning has lead to major breakthroughs in many recognition tasks as well as natural language processing~\cite{Lecun2015} thanks to efficient algorithms, a gigantic amount of available data, and powerful computing resources. Today's consumer products that we use on a daily basis such as smartphones and digital assistants are equipped with applications powered by deep neural networks (DNNs). However, training a successful DNN is not trivial. Algorithms used in training a DNN may be patented or have restricted licenses. Collecting and labeling data is costly, and GPU-accelerated computing hardware is also expensive. For example, the ImageNet dataset~\cite{ILSVRC15} contains about 1.2 million images, and training a DNN for an image classification model on such a dataset will take days and weeks even on GPU-accelerated machines. Therefore, production-level trained DNN models have great business value, and the need to protect models from copyright infringement is an urgent issue.

Moreover, conventional platforms for sharing models such as Model Zoo~\cite{jia2014caffe}, Azure AI Gallery~\cite{azureai}, and Tensor Hub~\cite{tensorhub} allow us to share DNN models for research and development purposes. A recent security assessment of on-device models from Android applications showed that many mobile applications fine-tuned pre-trained models from Tensor Hub~\cite{huang2021robustness}. Since models from such sharing platforms are widely used in real-world applications, it is necessary to properly credit DNN model owners to protect intellectual property (IP).

\RA{There are two aspects of IP protection for DNN models: access control and ownership verification.
The former focuses on protecting the functionality of DNN models from unauthorized access~\cite{chen2018protect, pyone2020training}, and the latter addresses ownership verification by taking inspiration from digital watermarking.
In this paper, we focus on ownership verification of DNN models.}
Researchers have proposed various model watermarking methods~\cite{2017-ICMR-Uchida, 2018-USENIX-Yossi,2018-ACCCS-Zhang,2018-Arxiv-Rouhani,2019-NIPS-Fan, 2019-MIPR-Sakazawa, 2020-NCA-Le}. However, most of the existing DNN watermarking methods are not robust against piracy attacks as described in~\cite{wang2019attacks,li2019piracy}.

Therefore, in this paper, we propose a DNN watermarking method that uses a block-wise image transformation with a secret key. The proposed method has been inspired by adversarial defenses~\cite{2020-ICIP-Maung, maung2021block}, which were in turn inspired by perceptual image encryption methods, which were proposed for privacy-preserving machine learning~\cite{2018-ICCETW-Tanaka, 2019-ICIP-Warit, 2019-Access-Warit, kawamura2020privacy} and encryption-then-compression systems~\cite{2019-TIFS-Chuman, 2019-APSIPAT-Warit, 2017-IEICE-Kurihara, chuman2017security}. The underlying idea of the proposed method is to embed watermark patterns in models by training the models with both plain images and transformed ones. Ownership is verified by matching the prediction of plain images and that of transformed ones. In experiments, the performance of protected models is close to that of non-protected ones, and the proposed method is also demonstrated to be robust against fine-tuning and model pruning attacks.

\section{Related Work\label{sec:related-work}}
\subsection{DNN Model Watermarking}
Inspired by digital watermarking, researchers have proposed various methods for preventing the illegal distribution of DNN models. There are mainly two approaches in DNN model watermarking: white-box and black-box.

White-box approaches require access to model weights for embedding and detecting watermarks in a DNN model. These methods use an embedding regularizer, which is an additional regularization term in a loss function during training~\cite{2017-ICMR-Uchida,nagai2018digital,2018-Arxiv-Chen, 2018-Arxiv-Rouhani}. A recent study~\cite{wang2019attacks} showed that these regularizer-based methods can be attacked. Another paper~\cite{2019-NIPS-Fan} highlighted that if watermarks are independent of a model's performance, they are vulnerable to ambiguity attacks~\cite{1998-IEEEJSAC-Craver} where two watermarks can be extracted from a protected model, causing confusion regarding ownership. Therefore, they introduced passports and passport layers~\cite{2019-NIPS-Fan}. However, a recent paper~\cite{li2019piracy} pointed out that ownership verification can be broken by using reverse-engineered secret passport weights. Accordingly, these white-box approaches are not practical in real-world applications such as online services because access to the model weights from a plagiarized party is not supported.

In black-box approaches, watermarks are extracted by observing the input and output of a model. A study in~\cite{2020-NCA-Le} introduced a black-box method by using adversarial examples. Another study in~\cite{2018-USENIX-Yossi} implanted a backdoor in a model so that a watermark can be triggered through the backdoor. Generally, in black-box approaches, a special set of training examples is used so that watermarks are extracted from the inference of a model~\cite{2018-ACCCS-Zhang,2019-NIPS-Fan, 2019-MIPR-Sakazawa, 2020-NCA-Le}. Li et al.\ pointed out that backdoor attack-based methods can be defeated by existing backdoor defenses (e.g.~\cite{wang2019neural}), and most of the existing methods are not robust enough against piracy attacks, where a verifiable watermark is injected into a model while maintaining the model's accuracy as described in~\cite{li2019piracy}.

Accordingly, we propose a DNN watermarking method that uses learnable transformed images with a secret key, in which the original watermark cannot be removed by piracy attacks. Similar to our work, Li et al.\ proposed a method called ``null embedding,'' which embeds a pattern into a model's decision process during the model's initial training~\cite{li2019piracy}. However, the effectiveness of their method has not been confirmed yet on large networks such as residual networks~\cite{2016-CVPR-He}, which are widely used for image classification tasks. In addition, the techniques used for transforming images are different; the proposed method uses a block-wise learnable transformation, in contrast to a null embedding pattern in~\cite{li2019piracy}.

\subsection{Learnable Image Encryption}
Learnable image encryption perceptually encrypts images while maintaining a network's ability to learn the encrypted ones for classification tasks. Most early methods of learnable image encryption were originally proposed to visually protect images for privacy-preserving DNNs~\cite{2018-ICCETW-Tanaka,madono2020block,2019-Access-Warit,2019-ICIP-Warit,sirichotedumrong2020gan,ito2020image,ito2020framework}.

Recently, adversarial defenses in~\cite{2020-ICIP-Maung,maung2021block} also utilized learnable image encryption methods. Instead of protecting visual information, these works focus on controlling a model's decision boundary with a secret key so that adversarial attacks are not effective on such models trained by learnable transformed images.

Another use case of learnable image encryption is the model protection proposed in~\cite{pyone2020training}. The study in~\cite{pyone2020training} focused on protecting a model from a functional perspective rather than ownership verification. In other words, a distributed model without a secret key is not usable. In this paper, a block-wise image transformation is applied to DNN watermarking for the first time.

\section{Threat Model\label{sec:threat}}
\begin{figure}[h]
 \centering
 \includegraphics[width=.8\linewidth]{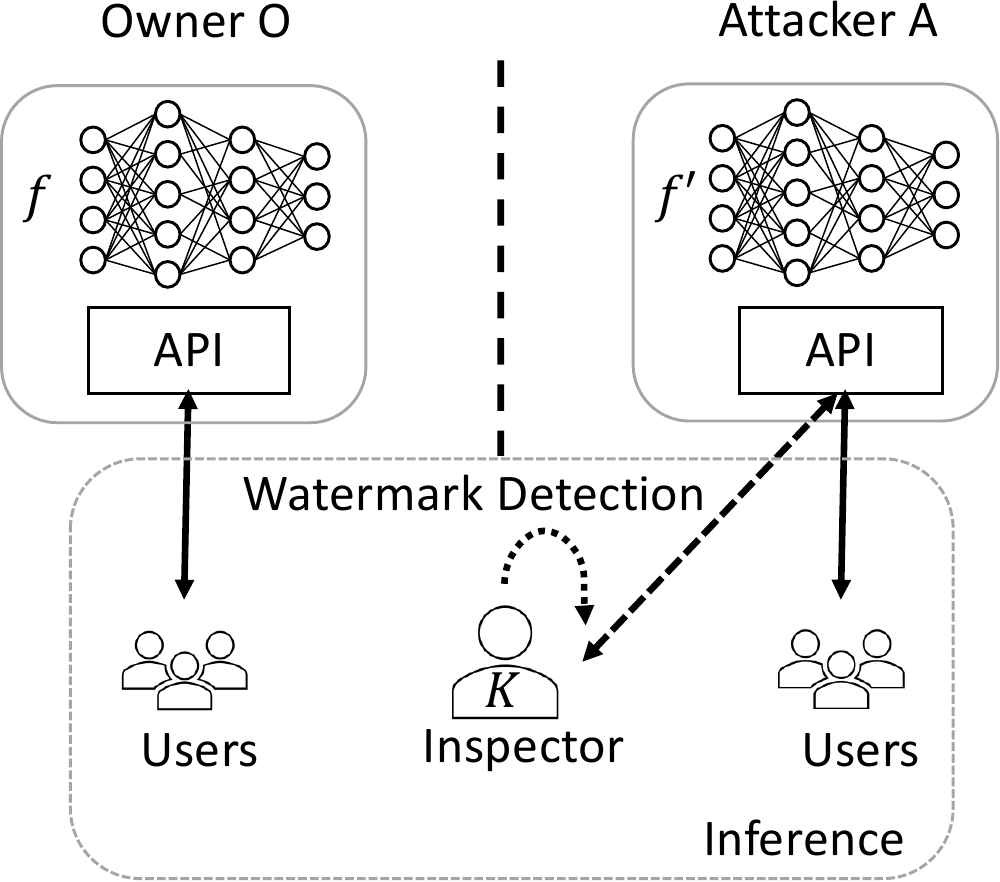}
 \caption{Application scenario of proposed DNN watermarking\label{fig:threat}}
\end{figure}

We consider an application scenario with two parties: owner $O$ and attacker $A$, as shown in Fig.~\ref{fig:threat}. Owner $O$ trains model $f$ with the proposed watermarking. Attacker $A$ illegally obtains model $f$ and establishes new model $f'$ with or without some modification to $f$. Both parties offer the same service via an application programming interface (API). When the model is in dispute, owner $O$ provides his/her secret key $K$ to an inspector, and the ownership is verified by using secret key $K$ through inference. We aim to verify the ownership of models by using secret key $K$ under this scenario.

There are two common ways of modifying models: pruning and fine-tuning. An attacker may use these methods to destroy watermarks in watermarked models.

\textbf{Fine-tuning:} Fine-tuning (transfer learning)~\cite{2015-ICLR-Simonyan} trains a model on top of pre-trained weights. Since fine-tuning alters the weights of a model, an attacker may use fine-tuning as an attack to overwrite a protected model with the intent of forging watermarks. We can consider such an attack scenario where the adversary has a subset of dataset $\mathcal{D}'$ and retrains the model with a forged key ($K'$).

\textbf{Pruning:} DNN models are often over-parameterized and contain millions of parameters. These giant models cannot be directly deployed on devices with limited resources such as smartphones, digital assistants, and embedded systems. Therefore, pruning techniques such as in~\cite{han2016eie,han2015deep,han2015learning,pavlo2017pruning} are used to compress the models by removing unimportant connections or neurons without losing accuracy. In this paper, parameter pruning is carried out by zeroing out weight values on the basis of the lowest L1-norm (i.e., to prune the weights that have the smallest absolute values) as in~\cite{2017-ICMR-Uchida}, and how it affects watermark detection is explored.

\section{Proposed DNN Watermarking\label{sec:proposed}}
\subsection{Overview}
\begin{figure}[h]
 \centering
 \includegraphics[width=\linewidth]{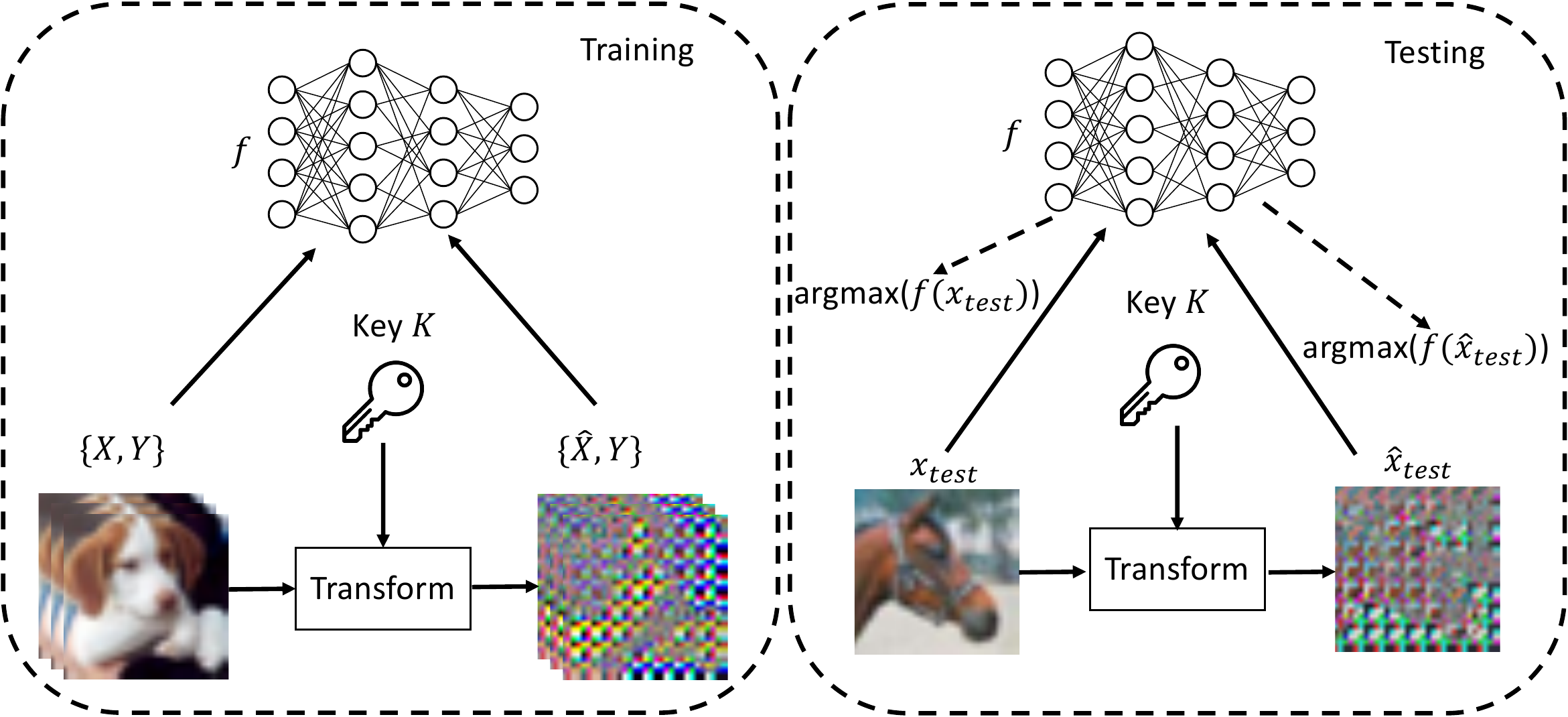}
 \caption{Overview of image classification with proposed DNN watermarking\label{fig:overview}}
\end{figure}

An overview of image classification with the proposed method is depicted in Fig.~\ref{fig:overview}. In the proposed DNN watermarking, model $f$ is trained with both clean images and images transformed by using secret key $K$. Such trained models are effective in classifying both plain images and transformed ones. This property enables us to verify the ownership of models. In addition, the watermark in the proposed watermarking cannot be removed, and adding a new watermark will decrease the model's accuracy. Therefore, the proposed method is piracy-resistant.

\subsection{Block-wise Transformation with Secret Key}
We use a block-wise negative/positive transformation with a secret key as in~\cite{maung2021block} to transform input images before training and validation of model ownership. The following are steps for transforming input images, where $c$, $w$, and $h$ denote the number of channels, width, and height of an image tensor $x \in {[0, 1]}^{c \times w \times h}$.

\begin{figure*}[h]
 \centering
 \subfloat[Original]{\includegraphics[width=0.16\linewidth]{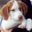}%
 \label{fig:dog}}
 \hfil
 \subfloat[$M = 2$]{\includegraphics[width=0.16\linewidth]{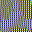}%
 \label{fig:np_2}}
 \hfil
 \subfloat[$M = 4$]{\includegraphics[width=0.16\linewidth]{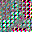}%
 \label{fig:np_4}}
 \hfil
 \subfloat[$M = 8$]{\includegraphics[width=0.16\linewidth]{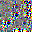}%
 \label{fig:np_8}}
 \hfil
 \subfloat[$M = 16$]{\includegraphics[width=0.16\linewidth]{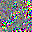}%
 \label{fig:np_16}}
 \hfil
 \subfloat[$M = 32$]{\includegraphics[width=0.16\linewidth]{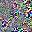}%
 \label{fig:np_32}}
 \caption{Example of block-wise transformed images\label{fig:images}}
\end{figure*}

\begin{enumerate}
\item Divide $x$ into blocks with a size of $M$ such that \\$\{B_{(1,1)}, \ldots, B_{(\frac{w}{M}, \frac{h}{M})}\}$.
\item Transform each block tensor $B_{(i, j)}$ into a vector \\$b_{(i,j)} = [b_{(i,j)}(1), \ldots, b_{(i,j)}(c \times M \times M)]$.
\item Generate key $K$, which is a binary vector, i.e.,
 \begin{equation}
 K = [K_1, \dots, K_k, \dots, K_{(c\times M \times M)}], K_k \in \{0, 1\},
 \end{equation}
where the value of the occurrence probability $P(K_k)$ is $0.5$.
\item Multiply each pixel value in $b_{(i, j)}$ by $255$ to be at $255$ scale with 8 bits.
\item Apply negative/positive transformation to every vector $b_{(i, j)}$ with $K$ as
 \begin{equation}
 b'_{(i, j)}(k) = \left\{
 \begin{array}{ll}
 b_{(i, j)}(k) & (K_k = 0)\\
 b_{(i, j)}(k) \oplus (2^L - 1) & (K_k = 1),
 \end{array}
 \right.
 \end{equation}
where $\oplus$ is an exclusive or (XOR) operation, $L$ is the number of bits used in $b_{(i, j)}(k)$, and $L = 8$ is used in this paper.
\item Divide each pixel value in $b'_{(i, j)}$ by $255$ to be at $[0, 1]$ scale.

\item Integrate the transformed vectors to form an image tensor $\hat{x} \in {[0, 1]}^{c \times w \times h}$.
\end{enumerate}
An example of images transformed by negative/positive transformation with different block sizes is shown in Fig.~\ref{fig:images}.

\subsection{Watermark Embedding}
A pattern caused by the transformation with key $K$ serves as a watermark in the proposed method. To embed the watermark in a DNN model, the model is trained by using transformed images. Let $X = \{x^1, \ldots, x^N\}$ be a set of training images and $Y = \{y^1,\ldots,y^N\}$ be a set of their respective truth labels in a one-hot vector. Algorithm~\ref{algo:embed} shows the watermark embedding process during training. Every image in $X$ is transformed with key $K$ to obtain a set of transformed images $\hat{X} = \{\hat{x}^1,\ldots,\hat{x}^N\}$. Model $f$ is trained by using both $X$ and $\hat{X}$.

\begin{algorithm}
\caption{Watermark Embedding\label{algo:embed}}
\begin{algorithmic}[1]
\renewcommand{\algorithmicrequire}{\textbf{Input:}}
\renewcommand{\algorithmicensure}{\textbf{Output:}}
\REQUIRE{$\{X, Y\}, K$}
\ENSURE{$f$}
\STATE{$\hat{X} \leftarrow$ \textsc{Transform} ($X, K$)}
\STATE{$f \leftarrow$ \textsc{Train} ($X, Y$)}
\STATE{$f \leftarrow$ \textsc{Train} ($\hat{X}, Y$)}
\end{algorithmic}
\end{algorithm}

\subsection{Watermark Detection}
To detect embedded watermarks, a statistical watermark-extraction method is used in the model inference. Let $X_{\text{test}} = \{x_{\text{test}}^1,\ldots,x_{\text{test}}^k,\\\ldots,x_{\text{test}}^s\}$ be a set of test images. Every image in $X_{\text{test}}$ is transformed with key $K$ to obtain $\hat{X}_{\text{test}} = \{\hat{x}_{\text{test}}^1,\ldots,\hat{x}_{\text{test}}^k,\ldots,\hat{x}_{\text{test}}^s\}$. Notably, $X_{\text{test}}$ is not a special pre-defined trigger set unlike conventional methods, so it can be a set of any test images within a classifier's distribution. In a typical image-classification scenario, $f$ takes a test image ($x_{\text{test}}^k$) and outputs a vector of unnormalized log probabilities (i.e., logits) as $f(x_{\text{test}}^k)$. In this paper, in accordance with this scenario, the class label of $x_{\text{test}}^k$ is estimated with the largest predicted probability, as $y_{\text{test}}^k = \text{argmax}(x_{\text{test}}^k)$.

Let $Y_{\text{label}} = \{y_{\text{test}}^1,\ldots,y_{\text{test}}^s\}$ be a set of predicted labels for $X_{\text{test}}$ and $\hat{Y}_{\text{label}} = \{\hat{y}_{\text{test}}^1,\ldots,\hat{y}_{\text{test}}^s\}$ be a set of predicted labels for $\hat{X}_{\text{test}}$. To evaluate the matching rate between $Y_{\text{label}}$ and $\hat{Y}_{\text{label}}$, the watermark detection accuracy $\tau$ is defined by
\begin{equation}
 \tau = \frac{1}{s}\sum_{k=1}^{s} \mathbbm{1} (y_{\text{test}}^k = \hat{y}_{\text{test}}^k), \label{eq:tau}
\end{equation}
where $s$ is the number of test images, and $\mathbbm{1}(\text{condition})$ is a value of one if the condition is satisfied, otherwise a value of zero.

To verify the ownership of a model, an inspector needs to set a threshold $th$. By using $th$, the watermark detection process is carried out as in Algorithm~\ref{algo:detect}. If $\tau$ is greater than $th$, the ownership verification is successful, and model $f$ is judged to be owner O's model.

\begin{algorithm}
\caption{Watermark Detection\label{algo:detect}}
\begin{algorithmic}[1]
\renewcommand{\algorithmicrequire}{\textbf{Input:}}
\renewcommand{\algorithmicensure}{\textbf{Output:}}
\REQUIRE{$f, X_{\text{test}}, K, th$}
\ENSURE{Successful or Unsuccessful}
\STATE{$\hat{X}_{\text{test}} \leftarrow$ \textsc{Transform} ($X_{\text{test}}, K$)}
\STATE{$\tau \leftarrow$ \textsc{Calculate\_Tau} ($f, X_{\text{test}}, \hat{X}_{\text{test}}$)}
\COMMENT{Equation~\ref{eq:tau}}
\IF{$\tau > th$}
\STATE{Successful}
\ELSE
\STATE{Unsuccessful}
\ENDIF
\end{algorithmic}
\end{algorithm}

\subsection{Properties of Proposed Method}
The proposed DNN watermarking method holds the following important properties:
\begin{itemize}
\item \textbf{Piracy-Resistance:} Original watermarks in a model cannot be removed, and adding new watermarks will decrease the model's accuracy.
\item \textbf{Low Computation Cost:} The block-wise operation can be efficiently implemented by using vectorized operations, and thus, \RA{pre-processing images with block-wise transformation in the proposed watermarking does not cause any noticeable overheads during training/inference.}
\item \textbf{Watermark Detection without a Trigger Set:} The proposed method uses a secret key to verify ownership. Therefore, a special trigger set with pre-defined labels for detecting a watermark is not required in the proposed method.
\end{itemize}

\section{Experiments\label{sec:experiments}}
\subsection{Setup\label{sec:setup}}
We conducted image classification experiments on the CIFAR-10 \\dataset~\cite{2009-Report-Krizhevsky} with a batch size of 128 and live augmentation (random cropping with padding of 4 and random horizontal flip) on a training set. CIFAR-10 consists of 60,000 color images (dimension of $32 \times 32 \times 3$) with 10 classes (6000 images for each class) where 50,000 images are for training and 10,000 for testing. We used deep residual networks~\cite{2016-CVPR-He} with 18 layers (ResNet18) and trained models for $200$ epochs with cyclic learning rates~\cite{2017-Arxiv-Smith} and mixed-precision training~\cite{2017-Arxiv-Micikevicius}. The parameters of the stochastic gradient descent (SGD) optimizer were a momentum of $0.9$, a weight decay of $0.0005$, and a maximum learning rate of $0.2$.

\subsection{Classification Performance and Watermark Detection}
We trained models by using the proposed method under five block sizes (i.e., $M \in \{2, 4, 8, 16, 32\}$). We evaluated the models in terms of classification accuracy (ACC) under three conditions: using plain images (plain), using transformed images with correct key $K$, and using transformed images with incorrect key $K'$. We also calculated the watermark detection accuracy (WDA) $\tau$ for correct key $K$ and WDA $\tau'$ for incorrect key $K'$.

\RA{Correct key $K$ was generated by using a random number generator from the PyTorch platform with a seed value of 42 (64-bit integer), and incorrect key $K'$ was also generated by using the same random number generator with a seed value of 123 (64-bit integer).} 

Table~\ref{tab:results} summarizes the results obtained under the above conditions. The models with a small block size such as $M = 2$ and $4$ performed better in detecting watermarks than that with $M = 8$, $16$, and $32$. The baseline model, which was a standard model trained by using plain images, was confirmed to have a low WDA because the model did not have a watermark. Since models with $M = 2$ and $4$ maintained a high classification accuracy when correct key $K$ was used, while the accuracy severely dropped when incorrect key $K'$ was given, we will focus on models with $M = 2$ and $4$ for further evaluation against attacks.

\robustify\bfseries
\sisetup{table-parse-only,detect-weight=true,detect-inline-weight=text,round-mode=places,round-precision=2}
\begin{table}
 \caption{Classification Accuracy (\SI{}{\percent}) and Watermark Detection Accuracy (\SI{}{\percent}) of Protected Models and Baseline Model. Values were averaged over testing whole test set (10,000 images).\label{tab:results}}
 \centering
 \begin{tabular}{l|S|SS|SS}
 \toprule
 & {ACC} & {ACC} & {WDA} & {ACC} & {WDA}\\
 {Model} & {(plain)} & {($K$)} & {($\tau$)} & {($K'$)} & {($\tau'$)}\\
 \midrule
 {$M = 2$} & 92.74 & 93.43 & 95.87 & 10.53 & 10.260\\
 {$M = 4$} & 92.99 & 92.24 & 94.20 & 15.55 & 15.75\\
 \midrule
 {$M = 8$} & 93.52 & 87.25 & 89.18 & 73.40 & 75.00\\
 {$M = 16$} & 93.71 & 89.26 & 90.50 & 82.21 & 83.87\\
 {$M = 32$} & 93.88 & 89.00 & 91.08 & 85.51 & 87.78\\
 \midrule
 {Baseline} & 95.45 & 11.34 & 11.43 & 12.02 & 12.12\\
 \bottomrule
 \end{tabular}
\end{table}

\subsection{Robustness Against Fine-tuning Attacks}
As described in the threat model (see Section~\ref{sec:threat}), we assumed an attacker obtains a small subset of training dataset $\mathcal{D'}$ ($\left| \mathcal{D}' \right| \in \{100, 500, 5000\}$). We fine-tuned the models with $M = 2$ and $4$ by using $\mathcal{D}'$ and new key $K'$ to embed a new watermark for 30 epochs with the same training settings as in Section~\ref{sec:setup} \RA{and Algorithm~\ref{algo:embed}.}

Table~\ref{tab:fine-tune} shows the results of fine-tuning attacks: model accuracies before and after fine-tuning, WDA $\tau$ for correct key $K$, and WDA $\tau'$ for new key $K'$. In any of the cases, fine-tuning attacks impaired the model accuracy, and WDA $\tau$ was greater than WDA $\tau'$. Therefore, the proposed method was confirmed to have resistance against piracy attacks.
\robustify\bfseries
\sisetup{table-parse-only,detect-weight=true,detect-inline-weight=text,round-mode=places,round-precision=2}
\begin{table*}
\centering
\caption{Classification Accuracy (\SI{}{\percent}) and Watermark Detection Accuracy (\SI{}{\percent}) of Protected Models Under Fine-Tuning Attacks. Values were averaged over testing whole test set (10,000 images).\label{tab:fine-tune}}
\begin{tabular}{l|SS|SSS|SSS|SSS}
 \toprule
& & & \multicolumn{3}{c|}{\circled{1} $\left| \mathcal{D}' \right| = 100$} & \multicolumn{3}{c|}{\circled{2} $\left| \mathcal{D}' \right| = 500$} & \multicolumn{3}{c}{\circled{3} $\left| \mathcal{D}' \right| = 5000$}\\
& {ACC} & {WDA} & {Fine-tuned} & {WDA} & {WDA} & {Fine-tuned} & {WDA} & {WDA} & {Fine-tuned} & {WDA} & {WDA}\\
 {Model} & {(plain)} & {($\tau$)} & {ACC} & {($\tau$)} & {($\tau'$)} & {ACC} & {($\tau$)} & {($\tau'$)} & {ACC} & {($\tau$)} & {($\tau'$)}\\
\midrule
 {$M = 2$} & 92.74 & 95.87 & 89.44 & 93.64 & 13.26 & 83.59 & 88.84 & 31.93 & 86.37 & 87.11 & 84.26\\
 {$M = 4$} & 92.99 & 94.20 & 91.79 & 93.46 & 16.50 & 87.50 & 89.90 & 23.14 & 82.62 & 71.24 & 69.15\\
\bottomrule
\end{tabular}
\end{table*}

\subsection{Robustness Against Pruning Attacks}
\begin{figure}[h]
 \centering
 \includegraphics[width=\linewidth]{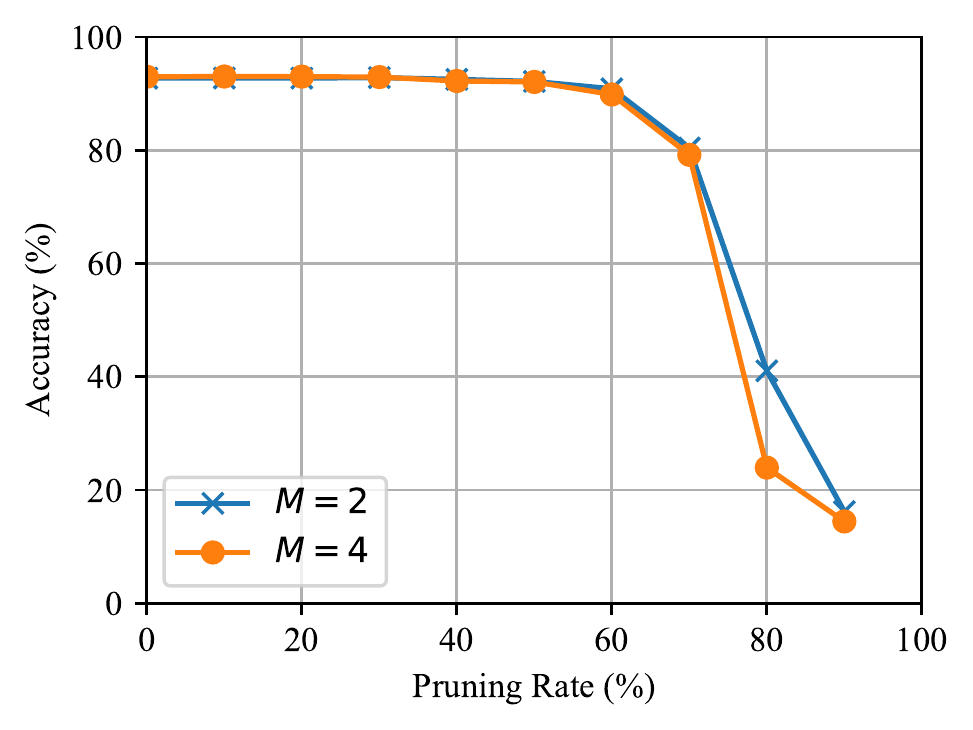}
 \caption{Classification accuracy under pruning attacks\label{fig:prune_acc}}
\end{figure}

We observed the classification accuracy and watermark detection accuracy $\tau$ under different pruning rates. Figure~\ref{fig:prune_acc} shows a graph of accuracy against pruning rates. The proposed method was robust against up to a pruning rate of \SI{60}{\percent}. After pruning more than \SI{60}{\percent}, both the accuracy and $\tau$ dropped.

\vspace{-2mm}

\begin{figure}[h]
 \centering
 \includegraphics[width=\linewidth]{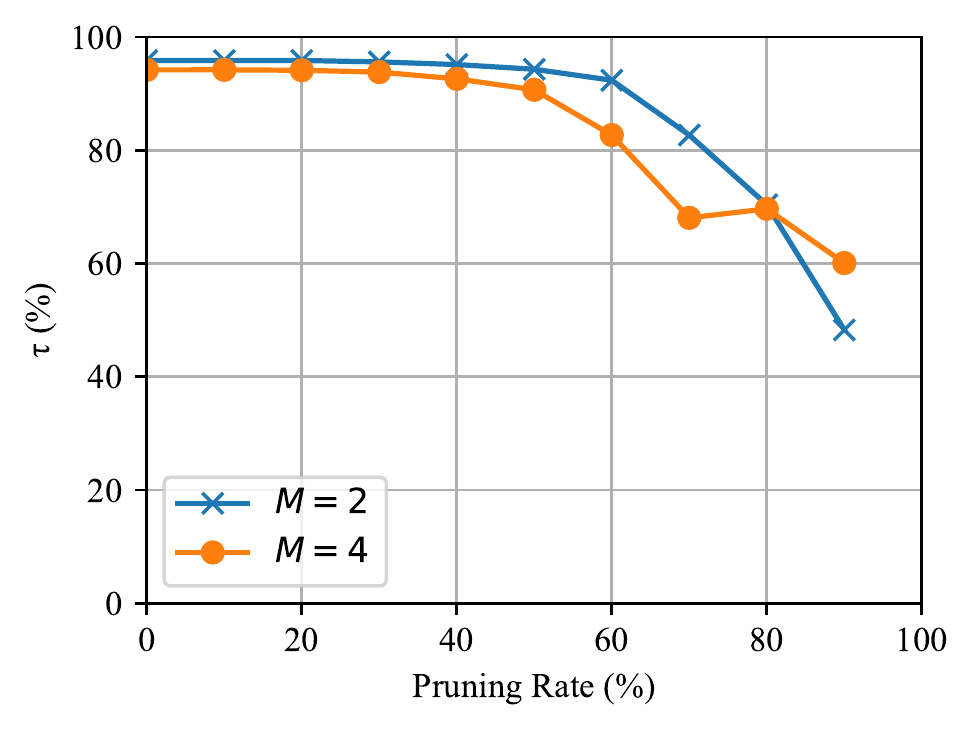}
 \caption{Watermark detection accuracy under pruning attacks\label{fig:prune_tau}}
\end{figure}
\vspace{-2mm}

\subsection{High-level Comparison with State-of-the-art Methods}
Table~\ref{tab:comparison} provides a high-level overview of state-of-the-art DNN watermarking methods in black-box settings. Embedding and verification methods vary from method to method. Most of the existing methods~\cite{2018-USENIX-Yossi,2020-NCA-Le,2018-ACCCS-Zhang,2019-NIPS-Fan} are not robust to piracy attacks as described in~\cite{li2019piracy}. In contrast, the watermark patterns used in the proposed method and Li et al.'s method~\cite{li2019piracy} are directly dependent on a model's accuracy. Therefore, piracy attacks will deteriorate a model's performance, and the original watermark detection will still be stronger than the pirated one. Note that the work in~\cite{li2019piracy} was evaluated only on a small convolutional network, and the proposed method was tested on a residual network with 18 layers (ResNet18). Therefore, the effectiveness of the proposed method was confirmed for practical scenarios.
\robustify\bfseries
\sisetup{table-parse-only,detect-weight=true,detect-inline-weight=text,round-mode=places,round-precision=2}
\begin{table*}
\centering
\caption{High-Level Comparison With State-of-the-Art Black-Box DNN Watermarking Methods\label{tab:comparison}}
\begin{tabular}{lccc}

\toprule
{Model} & {Embedding Method} & {Verification Method} & {Piracy Resistance}\\
\midrule
{Adi et al.~\cite{2018-USENIX-Yossi}} & {Backdoor} & {Trigger Set} & {No}\\
{Merrer et al.~\cite{2020-NCA-Le}} & {Adversarial Examples} & {Trigger Set} & {No}\\
{Zhang et al.~\cite{2018-ACCCS-Zhang}} & {Watermarked Examples} & {Trigger Set} & {No}\\
{Fan et al.~\cite{2019-NIPS-Fan}} & {Passport Layers + Trigger Set} & {Passports + Trigger Set} & {No}\\
{Li et al.~\cite{li2019piracy}}$^{\dagger}$ & {Null Embedding + Trigger Set} & {Watermark Accuracy + Trigger Set} & {Yes}\\
{Ours}$^{\ddagger}$ & {Learnable Image Transformation} & {Watermark Detection Accuracy} & {Yes}\\
\bottomrule
\multicolumn{4}{l}{$^{\dagger}$ Evaluated on a small convolutional neural network. $^{\ddagger}$ Evaluated on ResNet18.}\\
\end{tabular}
\end{table*}

\section{Conclusion\label{sec:conclusion}}
We proposed a novel model watermarking method that utilizes a learnable image transformation with a secret key for the first time. The proposed method trains a model by using both plain images and transformed ones and allows us to remotely verify the ownership of models. The results of experiments showed that the proposed method maintained a high classification accuracy, and watermarks in the proposed method could not be overwritten by piracy attacks. In addition, the proposed method was also robust against pruning attacks when parameters were pruned up to $\SI{60}{\percent}$.


\bibliographystyle{ACM-Reference-Format}
\bibliography{refs}


\begin{thebibliography}{44}


\ifx \showCODEN    \undefined \def \showCODEN     #1{\unskip}     \fi
\ifx \showDOI      \undefined \def \showDOI       #1{#1}\fi
\ifx \showISBNx    \undefined \def \showISBNx     #1{\unskip}     \fi
\ifx \showISBNxiii \undefined \def \showISBNxiii  #1{\unskip}     \fi
\ifx \showISSN     \undefined \def \showISSN      #1{\unskip}     \fi
\ifx \showLCCN     \undefined \def \showLCCN      #1{\unskip}     \fi
\ifx \shownote     \undefined \def \shownote      #1{#1}          \fi
\ifx \showarticletitle \undefined \def \showarticletitle #1{#1}   \fi
\ifx \showURL      \undefined \def \showURL       {\relax}        \fi
\providecommand\bibfield[2]{#2}
\providecommand\bibinfo[2]{#2}
\providecommand\natexlab[1]{#1}
\providecommand\showeprint[2][]{arXiv:#2}

\bibitem[\protect\citeauthoryear{??}{azu}{2021}]%
        {azureai}
 \bibinfo{year}{2021}\natexlab{}.
\newblock \bibinfo{title}{Azure AI Gallery}.
\newblock \bibinfo{howpublished}{\url{https://gallery.azure.ai/}}.
\newblock


\bibitem[\protect\citeauthoryear{??}{ten}{2021}]%
        {tensorhub}
 \bibinfo{year}{2021}\natexlab{}.
\newblock \bibinfo{title}{TensorFlow Hub is a repository of trained machine
  learning models}.
\newblock \bibinfo{howpublished}{\url{https://www.tensorflow.org/hub}}.
\newblock


\bibitem[\protect\citeauthoryear{Adi, Baum, Ciss{\'{e}}, Pinkas, and
  Keshet}{Adi et~al\mbox{.}}{2018}]%
        {2018-USENIX-Yossi}
\bibfield{author}{\bibinfo{person}{Yossi Adi}, \bibinfo{person}{Carsten Baum},
  \bibinfo{person}{Moustapha Ciss{\'{e}}}, \bibinfo{person}{Benny Pinkas},
  {and} \bibinfo{person}{Joseph Keshet}.} \bibinfo{year}{2018}\natexlab{}.
\newblock \showarticletitle{Turning Your Weakness Into a Strength: Watermarking
  Deep Neural Networks by Backdooring}. In \bibinfo{booktitle}{\emph{27th
  {USENIX} Security Symposium}}. \bibinfo{pages}{1615--1631}.
\newblock


\bibitem[\protect\citeauthoryear{AprilPyone and Kiya}{AprilPyone and
  Kiya}{2020a}]%
        {2020-ICIP-Maung}
\bibfield{author}{\bibinfo{person}{MaungMaung AprilPyone} {and}
  \bibinfo{person}{Hitoshi Kiya}.} \bibinfo{year}{2020}\natexlab{a}.
\newblock \showarticletitle{Encryption Inspired Adversarial Defense For Visual
  Classification}. In \bibinfo{booktitle}{\emph{2020 IEEE International
  Conference on Image Processing (ICIP)}}. \bibinfo{pages}{1681--1685}.
\newblock


\bibitem[\protect\citeauthoryear{AprilPyone and Kiya}{AprilPyone and
  Kiya}{2020b}]%
        {pyone2020training}
\bibfield{author}{\bibinfo{person}{MaungMaung AprilPyone} {and}
  \bibinfo{person}{Hitoshi Kiya}.} \bibinfo{year}{2020}\natexlab{b}.
\newblock \showarticletitle{Training DNN Model with Secret Key for Model
  Protection}. In \bibinfo{booktitle}{\emph{2020 IEEE 9th Global Conference on
  Consumer Electronics (GCCE)}}. \bibinfo{pages}{818--821}.
\newblock


\bibitem[\protect\citeauthoryear{AprilPyone and Kiya}{AprilPyone and
  Kiya}{2021}]%
        {maung2021block}
\bibfield{author}{\bibinfo{person}{MaungMaung AprilPyone} {and}
  \bibinfo{person}{Hitoshi Kiya}.} \bibinfo{year}{2021}\natexlab{}.
\newblock \showarticletitle{Block-wise Image Transformation with Secret Key for
  Adversarially Robust Defense}.
\newblock \bibinfo{journal}{\emph{IEEE Transactions on Information Forensics
  and Security}}  \bibinfo{volume}{16} (\bibinfo{year}{2021}),
  \bibinfo{pages}{2709--2723}.
\newblock


\bibitem[\protect\citeauthoryear{Chen, Rouhani, and Koushanfar}{Chen
  et~al\mbox{.}}{2018}]%
        {2018-Arxiv-Chen}
\bibfield{author}{\bibinfo{person}{Huili Chen}, \bibinfo{person}{Bita~Darvish
  Rouhani}, {and} \bibinfo{person}{Farinaz Koushanfar}.}
  \bibinfo{year}{2018}\natexlab{}.
\newblock \showarticletitle{DeepMarks: {A} Digital Fingerprinting Framework for
  Deep Neural Networks}.
\newblock \bibinfo{journal}{\emph{arXiv:1804.03648}} (\bibinfo{year}{2018}).
\newblock
\urldef\tempurl%
\url{http://arxiv.org/abs/1804.03648}
\showURL{%
\tempurl}


\bibitem[\protect\citeauthoryear{Chen and Wu}{Chen and Wu}{2018}]%
        {chen2018protect}
\bibfield{author}{\bibinfo{person}{Mingliang Chen} {and} \bibinfo{person}{Min
  Wu}.} \bibinfo{year}{2018}\natexlab{}.
\newblock \showarticletitle{Protect Your Deep Neural Networks from Piracy}. In
  \bibinfo{booktitle}{\emph{2018 IEEE International Workshop on Information
  Forensics and Security (WIFS)}}. IEEE, \bibinfo{pages}{1--7}.
\newblock


\bibitem[\protect\citeauthoryear{Chuman, Kurihara, and Kiya}{Chuman
  et~al\mbox{.}}{2017}]%
        {chuman2017security}
\bibfield{author}{\bibinfo{person}{Tatsuya Chuman}, \bibinfo{person}{Kenta
  Kurihara}, {and} \bibinfo{person}{Hitoshi Kiya}.}
  \bibinfo{year}{2017}\natexlab{}.
\newblock \showarticletitle{Security evaluation for block scrambling-based etc
  systems against extended jigsaw puzzle solver attacks}. In
  \bibinfo{booktitle}{\emph{2017 IEEE International Conference on Multimedia
  and Expo (ICME)}}. \bibinfo{pages}{229--234}.
\newblock


\bibitem[\protect\citeauthoryear{Chuman, Sirichotedumrong, and Kiya}{Chuman
  et~al\mbox{.}}{2019}]%
        {2019-TIFS-Chuman}
\bibfield{author}{\bibinfo{person}{Tatsuya Chuman}, \bibinfo{person}{Warit
  Sirichotedumrong}, {and} \bibinfo{person}{Hitoshi Kiya}.}
  \bibinfo{year}{2019}\natexlab{}.
\newblock \showarticletitle{Encryption-Then-Compression Systems Using
  Grayscale-Based Image Encryption for JPEG Images}.
\newblock \bibinfo{journal}{\emph{IEEE Transactions on Information Forensics
  and Security}} \bibinfo{volume}{14}, \bibinfo{number}{6}
  (\bibinfo{date}{June} \bibinfo{year}{2019}), \bibinfo{pages}{1515--1525}.
\newblock


\bibitem[\protect\citeauthoryear{Craver, Memon, Yeo, and Yeung}{Craver
  et~al\mbox{.}}{1998}]%
        {1998-IEEEJSAC-Craver}
\bibfield{author}{\bibinfo{person}{Scott Craver}, \bibinfo{person}{Nasir~D.
  Memon}, \bibinfo{person}{Boon{-}Lock Yeo}, {and} \bibinfo{person}{Minerva~M.
  Yeung}.} \bibinfo{year}{1998}\natexlab{}.
\newblock \showarticletitle{Resolving rightful ownerships with invisible
  watermarking techniques: limitations, attacks, and implications}.
\newblock \bibinfo{journal}{\emph{{IEEE} J. Sel. Areas Commun.}}
  \bibinfo{volume}{16}, \bibinfo{number}{4} (\bibinfo{year}{1998}),
  \bibinfo{pages}{573--586}.
\newblock


\bibitem[\protect\citeauthoryear{Fan, Ng, and Chan}{Fan et~al\mbox{.}}{2019}]%
        {2019-NIPS-Fan}
\bibfield{author}{\bibinfo{person}{Lixin Fan}, \bibinfo{person}{KamWoh Ng},
  {and} \bibinfo{person}{Chee~Seng Chan}.} \bibinfo{year}{2019}\natexlab{}.
\newblock \showarticletitle{Rethinking Deep Neural Network Ownership
  Verification: Embedding Passports to Defeat Ambiguity Attacks}. In
  \bibinfo{booktitle}{\emph{Advances in Neural Information Processing
  Systems}}. \bibinfo{pages}{4716--4725}.
\newblock


\bibitem[\protect\citeauthoryear{Han, Liu, Mao, Pu, Pedram, Horowitz, and
  Dally}{Han et~al\mbox{.}}{2016a}]%
        {han2016eie}
\bibfield{author}{\bibinfo{person}{Song Han}, \bibinfo{person}{Xingyu Liu},
  \bibinfo{person}{Huizi Mao}, \bibinfo{person}{Jing Pu},
  \bibinfo{person}{Ardavan Pedram}, \bibinfo{person}{Mark~A Horowitz}, {and}
  \bibinfo{person}{William~J Dally}.} \bibinfo{year}{2016}\natexlab{a}.
\newblock \showarticletitle{EIE: Efficient inference engine on compressed deep
  neural network}.
\newblock \bibinfo{journal}{\emph{ACM SIGARCH Computer Architecture News}}
  \bibinfo{volume}{44}, \bibinfo{number}{3} (\bibinfo{year}{2016}),
  \bibinfo{pages}{243--254}.
\newblock


\bibitem[\protect\citeauthoryear{Han, Mao, and Dally}{Han
  et~al\mbox{.}}{2016b}]%
        {han2015deep}
\bibfield{author}{\bibinfo{person}{Song Han}, \bibinfo{person}{Huizi Mao},
  {and} \bibinfo{person}{William~J. Dally}.} \bibinfo{year}{2016}\natexlab{b}.
\newblock \showarticletitle{Deep Compression: Compressing Deep Neural Network
  with Pruning, Trained Quantization and Huffman Coding}. In
  \bibinfo{booktitle}{\emph{International Conference on Learning
  Representations}}.
\newblock


\bibitem[\protect\citeauthoryear{Han, Pool, Tran, and Dally}{Han
  et~al\mbox{.}}{2015}]%
        {han2015learning}
\bibfield{author}{\bibinfo{person}{Song Han}, \bibinfo{person}{Jeff Pool},
  \bibinfo{person}{John Tran}, {and} \bibinfo{person}{William Dally}.}
  \bibinfo{year}{2015}\natexlab{}.
\newblock \showarticletitle{Learning both Weights and Connections for Efficient
  Neural Network}. In \bibinfo{booktitle}{\emph{Advances in Neural Information
  Processing Systems}}, \bibfield{editor}{\bibinfo{person}{C.~Cortes},
  \bibinfo{person}{N.~Lawrence}, \bibinfo{person}{D.~Lee},
  \bibinfo{person}{M.~Sugiyama}, {and} \bibinfo{person}{R.~Garnett}} (Eds.),
  Vol.~\bibinfo{volume}{28}. \bibinfo{publisher}{Curran Associates, Inc.},
  \bibinfo{pages}{1135--1143}.
\newblock


\bibitem[\protect\citeauthoryear{He, Zhang, Ren, and Sun}{He
  et~al\mbox{.}}{2016}]%
        {2016-CVPR-He}
\bibfield{author}{\bibinfo{person}{Kaiming He}, \bibinfo{person}{Xiangyu
  Zhang}, \bibinfo{person}{Shaoqing Ren}, {and} \bibinfo{person}{Jian Sun}.}
  \bibinfo{year}{2016}\natexlab{}.
\newblock \showarticletitle{Deep residual learning for image recognition}. In
  \bibinfo{booktitle}{\emph{Proceedings of the IEEE conference on computer
  vision and pattern recognition}}. \bibinfo{pages}{770--778}.
\newblock


\bibitem[\protect\citeauthoryear{Huang, Hu, and Chen}{Huang
  et~al\mbox{.}}{2021}]%
        {huang2021robustness}
\bibfield{author}{\bibinfo{person}{Yujin Huang}, \bibinfo{person}{Han Hu},
  {and} \bibinfo{person}{Chunyang Chen}.} \bibinfo{year}{2021}\natexlab{}.
\newblock \showarticletitle{Robustness of on-device Models: Adversarial Attack
  to Deep Learning Models on Android Apps}.
\newblock \bibinfo{journal}{\emph{arXiv:2101.04401}} (\bibinfo{year}{2021}).
\newblock
\urldef\tempurl%
\url{https://arxiv.org/abs/2101.04401}
\showURL{%
\tempurl}


\bibitem[\protect\citeauthoryear{Ito, Kinoshita, and Kiya}{Ito
  et~al\mbox{.}}{2020a}]%
        {ito2020framework}
\bibfield{author}{\bibinfo{person}{Hiroki Ito}, \bibinfo{person}{Yuma
  Kinoshita}, {and} \bibinfo{person}{Hitoshi Kiya}.}
  \bibinfo{year}{2020}\natexlab{a}.
\newblock \showarticletitle{A Framework for Transformation Network Training in
  Coordination with Semi-trusted Cloud Provider for Privacy-Preserving Deep
  Neural Networks}. In \bibinfo{booktitle}{\emph{2020 Asia-Pacific Signal and
  Information Processing Association Annual Summit and Conference (APSIPA
  ASC)}}. \bibinfo{pages}{1420--1424}.
\newblock


\bibitem[\protect\citeauthoryear{Ito, Kinoshita, and Kiya}{Ito
  et~al\mbox{.}}{2020b}]%
        {ito2020image}
\bibfield{author}{\bibinfo{person}{Hiroki Ito}, \bibinfo{person}{Yuma
  Kinoshita}, {and} \bibinfo{person}{Hitoshi Kiya}.}
  \bibinfo{year}{2020}\natexlab{b}.
\newblock \showarticletitle{Image transformation network for privacy-preserving
  deep neural networks and its security evaluation}. In
  \bibinfo{booktitle}{\emph{2020 IEEE 9th Global Conference on Consumer
  Electronics (GCCE)}}. IEEE, \bibinfo{pages}{822--825}.
\newblock


\bibitem[\protect\citeauthoryear{Jia, Shelhamer, Donahue, Karayev, Long,
  Girshick, Guadarrama, and Darrell}{Jia et~al\mbox{.}}{2014}]%
        {jia2014caffe}
\bibfield{author}{\bibinfo{person}{Yangqing Jia}, \bibinfo{person}{Evan
  Shelhamer}, \bibinfo{person}{Jeff Donahue}, \bibinfo{person}{Sergey Karayev},
  \bibinfo{person}{Jonathan Long}, \bibinfo{person}{Ross Girshick},
  \bibinfo{person}{Sergio Guadarrama}, {and} \bibinfo{person}{Trevor Darrell}.}
  \bibinfo{year}{2014}\natexlab{}.
\newblock \showarticletitle{Caffe: Convolutional architecture for fast feature
  embedding}. In \bibinfo{booktitle}{\emph{Proceedings of the 22nd ACM
  international conference on Multimedia}}. \bibinfo{pages}{675--678}.
\newblock


\bibitem[\protect\citeauthoryear{Kawamura, Kinoshita, Nakachi, Shiota, and
  Kiya}{Kawamura et~al\mbox{.}}{2020}]%
        {kawamura2020privacy}
\bibfield{author}{\bibinfo{person}{Ayana Kawamura}, \bibinfo{person}{Yuma
  Kinoshita}, \bibinfo{person}{Takayuki Nakachi}, \bibinfo{person}{Sayaka
  Shiota}, {and} \bibinfo{person}{Hitoshi Kiya}.}
  \bibinfo{year}{2020}\natexlab{}.
\newblock \showarticletitle{A privacy-preserving machine learning scheme using
  etc images}.
\newblock \bibinfo{journal}{\emph{IEICE Transactions on Fundamentals of
  Electronics, Communications and Computer Sciences}} \bibinfo{volume}{103},
  \bibinfo{number}{12} (\bibinfo{year}{2020}), \bibinfo{pages}{1571--1578}.
\newblock


\bibitem[\protect\citeauthoryear{Krizhevsky and Hinton}{Krizhevsky and
  Hinton}{2009}]%
        {2009-Report-Krizhevsky}
\bibfield{author}{\bibinfo{person}{Alex Krizhevsky} {and}
  \bibinfo{person}{Geoffrey Hinton}.} \bibinfo{year}{2009}\natexlab{}.
\newblock \bibinfo{booktitle}{\emph{Learning multiple layers of features from
  tiny images}}.
\newblock \bibinfo{type}{{T}echnical {R}eport}.
  \bibinfo{institution}{University of Toronto}.
\newblock


\bibitem[\protect\citeauthoryear{Kurihara, Imaizumi, Shiota, and Kiya}{Kurihara
  et~al\mbox{.}}{2017}]%
        {2017-IEICE-Kurihara}
\bibfield{author}{\bibinfo{person}{Kenta Kurihara}, \bibinfo{person}{Shoko
  Imaizumi}, \bibinfo{person}{Sayaka Shiota}, {and} \bibinfo{person}{Hitoshi
  Kiya}.} \bibinfo{year}{2017}\natexlab{}.
\newblock \showarticletitle{An encryption-then-compression system for lossless
  image compression standards}.
\newblock \bibinfo{journal}{\emph{IEICE transactions on information and
  systems}} \bibinfo{volume}{100}, \bibinfo{number}{1} (\bibinfo{year}{2017}),
  \bibinfo{pages}{52--56}.
\newblock


\bibitem[\protect\citeauthoryear{LeCun, Bengio, and Hinton}{LeCun
  et~al\mbox{.}}{2015}]%
        {Lecun2015}
\bibfield{author}{\bibinfo{person}{Yann LeCun}, \bibinfo{person}{Yoshua
  Bengio}, {and} \bibinfo{person}{Geoffrey Hinton}.}
  \bibinfo{year}{2015}\natexlab{}.
\newblock \showarticletitle{Deep learning}.
\newblock \bibinfo{journal}{\emph{nature}} \bibinfo{volume}{521},
  \bibinfo{number}{7553} (\bibinfo{year}{2015}), \bibinfo{pages}{436}.
\newblock


\bibitem[\protect\citeauthoryear{Li, Wenger, Zhao, and Zheng}{Li
  et~al\mbox{.}}{2019}]%
        {li2019piracy}
\bibfield{author}{\bibinfo{person}{Huiying Li}, \bibinfo{person}{Emily Wenger},
  \bibinfo{person}{Ben~Y Zhao}, {and} \bibinfo{person}{Haitao Zheng}.}
  \bibinfo{year}{2019}\natexlab{}.
\newblock \showarticletitle{Piracy resistant watermarks for deep neural
  networks}.
\newblock \bibinfo{journal}{\emph{arXiv:1910.01226}} (\bibinfo{year}{2019}).
\newblock
\urldef\tempurl%
\url{https://arxiv.org/abs/1910.01226}
\showURL{%
\tempurl}


\bibitem[\protect\citeauthoryear{Madono, Tanaka, Onishi, and Ogawa}{Madono
  et~al\mbox{.}}{2020}]%
        {madono2020block}
\bibfield{author}{\bibinfo{person}{Koki Madono}, \bibinfo{person}{Masayuki
  Tanaka}, \bibinfo{person}{Masaki Onishi}, {and} \bibinfo{person}{Tetsuji
  Ogawa}.} \bibinfo{year}{2020}\natexlab{}.
\newblock \showarticletitle{Block-wise Scrambled Image Recognition Using
  Adaptation Network}.
\newblock \bibinfo{journal}{\emph{arXiv:2001.07761}} (\bibinfo{year}{2020}).
\newblock
\urldef\tempurl%
\url{https://arxiv.org/abs/2001.07761}
\showURL{%
\tempurl}


\bibitem[\protect\citeauthoryear{Merrer, P{\'{e}}rez, and Tr{\'{e}}dan}{Merrer
  et~al\mbox{.}}{2020}]%
        {2020-NCA-Le}
\bibfield{author}{\bibinfo{person}{Erwan~Le Merrer}, \bibinfo{person}{Patrick
  P{\'{e}}rez}, {and} \bibinfo{person}{Gilles Tr{\'{e}}dan}.}
  \bibinfo{year}{2020}\natexlab{}.
\newblock \showarticletitle{Adversarial frontier stitching for remote neural
  network watermarking}.
\newblock \bibinfo{journal}{\emph{Neural Computing and Applications}}
  \bibinfo{volume}{32}, \bibinfo{number}{13} (\bibinfo{year}{2020}),
  \bibinfo{pages}{9233--9244}.
\newblock


\bibitem[\protect\citeauthoryear{Micikevicius, Narang, Alben, Diamos, Elsen,
  Garc{\'{\i}}a, Ginsburg, Houston, Kuchaiev, Venkatesh, and Wu}{Micikevicius
  et~al\mbox{.}}{2017}]%
        {2017-Arxiv-Micikevicius}
\bibfield{author}{\bibinfo{person}{Paulius Micikevicius},
  \bibinfo{person}{Sharan Narang}, \bibinfo{person}{Jonah Alben},
  \bibinfo{person}{Gregory~F. Diamos}, \bibinfo{person}{Erich Elsen},
  \bibinfo{person}{David Garc{\'{\i}}a}, \bibinfo{person}{Boris Ginsburg},
  \bibinfo{person}{Michael Houston}, \bibinfo{person}{Oleksii Kuchaiev},
  \bibinfo{person}{Ganesh Venkatesh}, {and} \bibinfo{person}{Hao Wu}.}
  \bibinfo{year}{2017}\natexlab{}.
\newblock \showarticletitle{Mixed Precision Training}.
\newblock \bibinfo{journal}{\emph{arXiv:1710.03740}} (\bibinfo{year}{2017}).
\newblock
\urldef\tempurl%
\url{http://arxiv.org/abs/1710.03740}
\showURL{%
\tempurl}


\bibitem[\protect\citeauthoryear{Molchanov, Tyree, Karras, Aila, and
  Kautz}{Molchanov et~al\mbox{.}}{2017}]%
        {pavlo2017pruning}
\bibfield{author}{\bibinfo{person}{Pavlo Molchanov}, \bibinfo{person}{Stephen
  Tyree}, \bibinfo{person}{Tero Karras}, \bibinfo{person}{Timo Aila}, {and}
  \bibinfo{person}{Jan Kautz}.} \bibinfo{year}{2017}\natexlab{}.
\newblock \showarticletitle{Pruning Convolutional Neural Networks for Resource
  Efficient Inference}. In \bibinfo{booktitle}{\emph{International Conference
  on Learning Representations}}.
\newblock


\bibitem[\protect\citeauthoryear{Nagai, Uchida, Sakazawa, and Satoh}{Nagai
  et~al\mbox{.}}{2018}]%
        {nagai2018digital}
\bibfield{author}{\bibinfo{person}{Yuki Nagai}, \bibinfo{person}{Yusuke
  Uchida}, \bibinfo{person}{Shigeyuki Sakazawa}, {and}
  \bibinfo{person}{Shin’ichi Satoh}.} \bibinfo{year}{2018}\natexlab{}.
\newblock \showarticletitle{Digital watermarking for deep neural networks}.
\newblock \bibinfo{journal}{\emph{International Journal of Multimedia
  Information Retrieval}} \bibinfo{volume}{7}, \bibinfo{number}{1}
  (\bibinfo{year}{2018}), \bibinfo{pages}{3--16}.
\newblock


\bibitem[\protect\citeauthoryear{Rouhani, Chen, and Koushanfar}{Rouhani
  et~al\mbox{.}}{2018}]%
        {2018-Arxiv-Rouhani}
\bibfield{author}{\bibinfo{person}{Bita~Darvish Rouhani},
  \bibinfo{person}{Huili Chen}, {and} \bibinfo{person}{Farinaz Koushanfar}.}
  \bibinfo{year}{2018}\natexlab{}.
\newblock \showarticletitle{DeepSigns: {A} Generic Watermarking Framework for
  {IP} Protection of Deep Learning Models}.
\newblock \bibinfo{journal}{\emph{arXiv:1804.00750}} (\bibinfo{year}{2018}).
\newblock
\urldef\tempurl%
\url{http://arxiv.org/abs/1804.00750}
\showURL{%
\tempurl}


\bibitem[\protect\citeauthoryear{Russakovsky, Deng, Su, Krause, Satheesh, Ma,
  Huang, Karpathy, Khosla, Bernstein, Berg, and Fei-Fei}{Russakovsky
  et~al\mbox{.}}{2015}]%
        {ILSVRC15}
\bibfield{author}{\bibinfo{person}{Olga Russakovsky}, \bibinfo{person}{Jia
  Deng}, \bibinfo{person}{Hao Su}, \bibinfo{person}{Jonathan Krause},
  \bibinfo{person}{Sanjeev Satheesh}, \bibinfo{person}{Sean Ma},
  \bibinfo{person}{Zhiheng Huang}, \bibinfo{person}{Andrej Karpathy},
  \bibinfo{person}{Aditya Khosla}, \bibinfo{person}{Michael Bernstein},
  \bibinfo{person}{Alexander~C. Berg}, {and} \bibinfo{person}{Li Fei-Fei}.}
  \bibinfo{year}{2015}\natexlab{}.
\newblock \showarticletitle{{ImageNet Large Scale Visual Recognition
  Challenge}}.
\newblock \bibinfo{journal}{\emph{International Journal of Computer Vision
  (IJCV)}} \bibinfo{volume}{115}, \bibinfo{number}{3} (\bibinfo{year}{2015}),
  \bibinfo{pages}{211--252}.
\newblock


\bibitem[\protect\citeauthoryear{Sakazawa, Myodo, Tasaka, and
  Yanagihara}{Sakazawa et~al\mbox{.}}{2019}]%
        {2019-MIPR-Sakazawa}
\bibfield{author}{\bibinfo{person}{Shigeyuki Sakazawa}, \bibinfo{person}{Emi
  Myodo}, \bibinfo{person}{Kazuyuki Tasaka}, {and} \bibinfo{person}{Hiromasa
  Yanagihara}.} \bibinfo{year}{2019}\natexlab{}.
\newblock \showarticletitle{Visual Decoding of Hidden Watermark in Trained Deep
  Neural Network}. In \bibinfo{booktitle}{\emph{2nd {IEEE} Conference on
  Multimedia Information Processing and Retrieval}}. \bibinfo{pages}{371--374}.
\newblock


\bibitem[\protect\citeauthoryear{Simonyan and Zisserman}{Simonyan and
  Zisserman}{2015}]%
        {2015-ICLR-Simonyan}
\bibfield{author}{\bibinfo{person}{Karen Simonyan} {and}
  \bibinfo{person}{Andrew Zisserman}.} \bibinfo{year}{2015}\natexlab{}.
\newblock \showarticletitle{Very Deep Convolutional Networks for Large-Scale
  Image Recognition}. In \bibinfo{booktitle}{\emph{International Conference on
  Learning Representations}}.
\newblock


\bibitem[\protect\citeauthoryear{Sirichotedumrong, Kinoshita, and
  Kiya}{Sirichotedumrong et~al\mbox{.}}{2019a}]%
        {2019-Access-Warit}
\bibfield{author}{\bibinfo{person}{Warit Sirichotedumrong},
  \bibinfo{person}{Yuma Kinoshita}, {and} \bibinfo{person}{Hitoshi Kiya}.}
  \bibinfo{year}{2019}\natexlab{a}.
\newblock \showarticletitle{Pixel-Based Image Encryption Without Key Management
  for Privacy-Preserving Deep Neural Networks}.
\newblock \bibinfo{journal}{\emph{IEEE Access}}  \bibinfo{volume}{7}
  (\bibinfo{year}{2019}), \bibinfo{pages}{177844--177855}.
\newblock


\bibitem[\protect\citeauthoryear{Sirichotedumrong and Kiya}{Sirichotedumrong
  and Kiya}{2019}]%
        {2019-APSIPAT-Warit}
\bibfield{author}{\bibinfo{person}{Warit Sirichotedumrong} {and}
  \bibinfo{person}{Hitoshi Kiya}.} \bibinfo{year}{2019}\natexlab{}.
\newblock \showarticletitle{Grayscale-based block scrambling image encryption
  using ycbcr color space for encryption-then-compression systems}.
\newblock \bibinfo{journal}{\emph{APSIPA Transactions on Signal and Information
  Processing}}  \bibinfo{volume}{8} (\bibinfo{year}{2019}).
\newblock


\bibitem[\protect\citeauthoryear{Sirichotedumrong and Kiya}{Sirichotedumrong
  and Kiya}{2020}]%
        {sirichotedumrong2020gan}
\bibfield{author}{\bibinfo{person}{Warit Sirichotedumrong} {and}
  \bibinfo{person}{Hitoshi Kiya}.} \bibinfo{year}{2020}\natexlab{}.
\newblock \showarticletitle{A GAN-Based Image Transformation Scheme for
  Privacy-Preserving Deep Neural Networks}. In \bibinfo{booktitle}{\emph{2020
  28th European Signal Processing Conference (EUSIPCO)}}.
  \bibinfo{pages}{745--749}.
\newblock


\bibitem[\protect\citeauthoryear{Sirichotedumrong, Maekawa, Kinoshita, and
  Kiya}{Sirichotedumrong et~al\mbox{.}}{2019b}]%
        {2019-ICIP-Warit}
\bibfield{author}{\bibinfo{person}{Warit Sirichotedumrong},
  \bibinfo{person}{Takahiro Maekawa}, \bibinfo{person}{Yuma Kinoshita}, {and}
  \bibinfo{person}{Hitoshi Kiya}.} \bibinfo{year}{2019}\natexlab{b}.
\newblock \showarticletitle{Privacy-preserving deep neural networks with
  pixel-based image encryption considering data augmentation in the encrypted
  domain}. In \bibinfo{booktitle}{\emph{2019 IEEE International Conference on
  Image Processing (ICIP)}}. \bibinfo{pages}{674--678}.
\newblock


\bibitem[\protect\citeauthoryear{Smith and Topin}{Smith and Topin}{2017}]%
        {2017-Arxiv-Smith}
\bibfield{author}{\bibinfo{person}{Leslie~N. Smith} {and}
  \bibinfo{person}{Nicholay Topin}.} \bibinfo{year}{2017}\natexlab{}.
\newblock \showarticletitle{Super-Convergence: Very Fast Training of Residual
  Networks Using Large Learning Rates}.
\newblock \bibinfo{journal}{\emph{arXiv:1708.07120}} (\bibinfo{year}{2017}).
\newblock
\urldef\tempurl%
\url{http://arxiv.org/abs/1708.07120}
\showURL{%
\tempurl}


\bibitem[\protect\citeauthoryear{Tanaka}{Tanaka}{2018}]%
        {2018-ICCETW-Tanaka}
\bibfield{author}{\bibinfo{person}{Masayuki Tanaka}.}
  \bibinfo{year}{2018}\natexlab{}.
\newblock \showarticletitle{Learnable Image Encryption}. In
  \bibinfo{booktitle}{\emph{2018 IEEE International Conference on Consumer
  Electronics-Taiwan (ICCE-TW)}}. \bibinfo{pages}{1--2}.
\newblock


\bibitem[\protect\citeauthoryear{Uchida, Nagai, Sakazawa, and Satoh}{Uchida
  et~al\mbox{.}}{2017}]%
        {2017-ICMR-Uchida}
\bibfield{author}{\bibinfo{person}{Yusuke Uchida}, \bibinfo{person}{Yuki
  Nagai}, \bibinfo{person}{Shigeyuki Sakazawa}, {and}
  \bibinfo{person}{Shin'ichi Satoh}.} \bibinfo{year}{2017}\natexlab{}.
\newblock \showarticletitle{Embedding Watermarks into Deep Neural Networks}. In
  \bibinfo{booktitle}{\emph{Proceedings of the 2017 {ACM} on International
  Conference on Multimedia Retrieval}}. \bibinfo{pages}{269--277}.
\newblock


\bibitem[\protect\citeauthoryear{Wang, Yao, Shan, Li, Viswanath, Zheng, and
  Zhao}{Wang et~al\mbox{.}}{2019}]%
        {wang2019neural}
\bibfield{author}{\bibinfo{person}{Bolun Wang}, \bibinfo{person}{Yuanshun Yao},
  \bibinfo{person}{Shawn Shan}, \bibinfo{person}{Huiying Li},
  \bibinfo{person}{Bimal Viswanath}, \bibinfo{person}{Haitao Zheng}, {and}
  \bibinfo{person}{Ben~Y Zhao}.} \bibinfo{year}{2019}\natexlab{}.
\newblock \showarticletitle{Neural cleanse: Identifying and mitigating backdoor
  attacks in neural networks}. In \bibinfo{booktitle}{\emph{2019 IEEE Symposium
  on Security and Privacy (SP)}}. IEEE, \bibinfo{pages}{707--723}.
\newblock


\bibitem[\protect\citeauthoryear{Wang and Kerschbaum}{Wang and
  Kerschbaum}{2019}]%
        {wang2019attacks}
\bibfield{author}{\bibinfo{person}{Tianhao Wang} {and} \bibinfo{person}{Florian
  Kerschbaum}.} \bibinfo{year}{2019}\natexlab{}.
\newblock \showarticletitle{Attacks on digital watermarks for deep neural
  networks}. In \bibinfo{booktitle}{\emph{ICASSP 2019-2019 IEEE International
  Conference on Acoustics, Speech and Signal Processing (ICASSP)}}. IEEE,
  \bibinfo{pages}{2622--2626}.
\newblock


\bibitem[\protect\citeauthoryear{Zhang, Gu, Jang, Wu, Stoecklin, Huang, and
  Molloy}{Zhang et~al\mbox{.}}{2018}]%
        {2018-ACCCS-Zhang}
\bibfield{author}{\bibinfo{person}{Jialong Zhang}, \bibinfo{person}{Zhongshu
  Gu}, \bibinfo{person}{Jiyong Jang}, \bibinfo{person}{Hui Wu},
  \bibinfo{person}{Marc~Ph. Stoecklin}, \bibinfo{person}{Heqing Huang}, {and}
  \bibinfo{person}{Ian Molloy}.} \bibinfo{year}{2018}\natexlab{}.
\newblock \showarticletitle{Protecting Intellectual Property of Deep Neural
  Networks with Watermarking}. In \bibinfo{booktitle}{\emph{Proceedings of the
  2018 on Asia Conference on Computer and Communications Security}}.
  \bibinfo{pages}{159--172}.
\newblock


\end{thebibliography}

\end{document}